\PassOptionsToPackage{table}{xcolor}
\documentclass[10pt,twocolumn,letterpaper]{article}
\usepackage[T1]{fontenc}
\usepackage[pagenumbers]{cvpr}

\usepackage{amsmath,amsfonts,bm}

\def\eqref#1{equation~\ref{#1}}

\def\1{\bm{1}}

\DeclareMathAlphabet{\mathsfit}{\encodingdefault}{\sfdefault}{m}{sl}
\SetMathAlphabet{\mathsfit}{bold}{\encodingdefault}{\sfdefault}{bx}{n}

\usepackage{microtype}
\usepackage{multirow}
\usepackage{newunicodechar}
\newunicodechar{’}{\textquoteright}
\usepackage{tikz}
\usetikzlibrary{arrows.meta,fit,positioning}

\definecolor{metricgray}{RGB}{245,245,245}
\definecolor{oursblue}{RGB}{228,240,252}
\definecolor{referencegray}{RGB}{128,128,128}

\definecolor{cbBlue}{HTML}{0072B2}
\definecolor{cbSky}{HTML}{56B4E9}
\definecolor{cbGreen}{HTML}{009E73}
\definecolor{cbOrange}{HTML}{E69F00}
\definecolor{cbPurple}{HTML}{CC79A7}
\definecolor{cbRed}{HTML}{D55E00}
\definecolor{cbYellow}{HTML}{F0E442}
\definecolor{cbGray}{HTML}{BDBDBD}

\newcommand{\method}{TrajWAM}
\newcommand{\sg}{\operatorname{sg}}

\newcommand{\baseplanner}[1][~]{Base Planner#1(Phase I)}
\newcommand{\planningft}[1][~]{Base Planner#1(planning-only FT)}
\newcommand{\futureft}[1][~]{Base Planner#1(unconditioned future FT)}
\newcommand{\nosepmodel}{\method{} w/o separation}
\newcommand{\fullmodel}{\method{}}

\definecolor{cvprblue}{rgb}{0.21,0.49,0.74}
\usepackage[breaklinks,colorlinks,allcolors=cvprblue]{hyperref}
\hypersetup{pdftitle={TrajWAM: Trajectory-Conditioned World Action Modeling for Vision--Language Driving},pdfauthor={Xuerui Fang, Gulimila Kezierbieke, CHEN WANGXING, Hui Li}}
\title{TrajWAM: Trajectory-Conditioned\\World Action Modeling for Vision--Language Driving}
\author{%
Xuerui Fang$^{1}$\quad Gulimila Kezierbieke$^{1,*}$\quad CHEN WANGXING$^{2}$\quad Hui Li$^{3,*}$\\
$^{1}$Xinjiang Agricultural University\\
$^{2}$ToReality (Hangzhou) Network Technology Co., Ltd.\\
$^{3}$Jiangnan University\\[0.5ex]
{\small $^{*}$Corresponding authors: \texttt{glml@xjau.edu.cn}, \texttt{lihui.cv@jiangnan.edu.cn}}
}

\begin{document}
\raggedbottom
\maketitle
\begin{center}
\small Code and models: \url{https://github.com/FangWD520/TrajWAM}
\end{center}

\begin{abstract}
World Action Models (WAMs) connect future scene prediction with action planning for autonomous driving. Driving records pair executed trajectories with observed futures, leaving alternative trajectory queries unlabeled. We present \method{}, a trajectory-conditioned WAM that improves vision--language driving through auxiliary bird's-eye-view (BEV) future learning. An ego-trajectory query conditions DINOv3-aligned BEV predictions from shared planner representations. Training combines factual reconstruction under the executed trajectory with one-sided feature separation under an unlabeled perturbed trajectory. Joint optimization with waypoint supervision transfers these auxiliary learning signals to the shared planner. Native inference directly generates waypoints and bypasses the auxiliary future-prediction branch. On all 220 Bench2Drive routes, \method{} achieves 91.53 Driving Score and 75.45\% Success Rate, exceeding a schedule-matched planning-only fine-tuning baseline by 5.31 Driving Score points and nine successful routes.

\end{abstract}

	\section{Introduction}
	\label{sec:intro}

	\begin{figure*}[t]
		\centering
		\includegraphics[width=\textwidth]{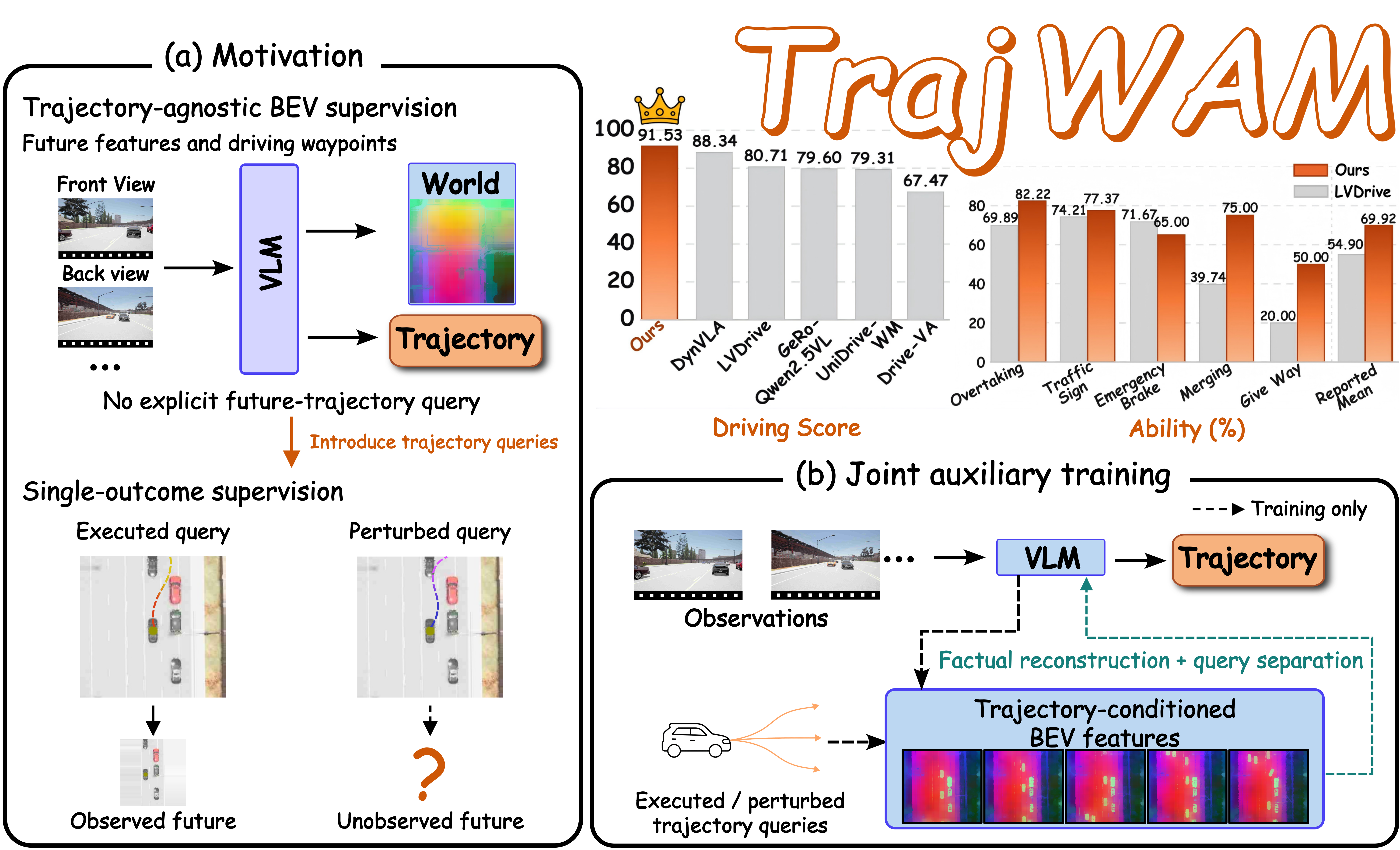}
		\caption{Motivation and overview of \method{}. (a) Trajectory-agnostic future supervision predicts future features and driving waypoints from shared scene representations. Introducing trajectory queries creates a supervision challenge: the executed query has an observed future, while alternative queries remain unlabeled. (b) Factual reconstruction and query separation jointly train the driving policy, which generates waypoints directly at inference. Upper right: reported Driving Score and ability comparisons.}
		\label{fig:motivation}
	\end{figure*}

	Autonomous driving requires recognizing the current scene and anticipating its evolution under ego actions. World Action Models (WAMs) connect observations and actions with future scene representations~\citep{wang2026wamsurvey,ye2026dreamzero}. Driving systems learn video, latent, or bird's-eye-view (BEV) futures to support planning~\citep{wang2024drivewm,li2025law,li2025wote,zheng2025world4drive}. When future prediction and planning share parameters, observed futures also supervise the driving policy during training. This coupling supports direct waypoint generation at deployment.

Representation-space supervision provides a compact interface for this coupling. DeepSight supervises DINOv3-aligned BEV futures within a driving VLM~\citep{zhang2026deepsight}, and Latent-WAM trains auxiliary latent dynamics alongside a driving policy~\citep{wang2026latentwam}. We study how an explicit future-trajectory query can strengthen this auxiliary supervision.

Trajectory-agnostic supervision predicts future features from scene states. A future-waypoint query explicitly links these predictions to ego motion (Figure~\ref{fig:motivation}(a)). The resulting training problem has asymmetric supervision: driving records supply the observed future for the executed trajectory, while alternative queries have unobserved outcomes. Our central question is: \emph{How can trajectory queries and observed futures jointly improve a vision--language driving policy?}

We introduce \method{}, a trajectory-conditioned WAM that trains a Qwen2.5-VL-3B driving planner with auxiliary BEV future supervision~\citep{bai2025qwen25vl}. A four-waypoint ego trajectory modulates the shared VLM's future-query states through feature-wise linear modulation (FiLM)~\citep{perez2018film}. A frozen readout maps these states to DINOv3-aligned features at the current anchor and four future times~\citep{simeoni2025dinov3}. The trajectory enters the readout after planner-logit computation, and the auxiliary losses backpropagate into shared planner layers. Native inference retains the fixed future-query tokens and directly generates reasoning and waypoints.

Our joint objective combines factual reconstruction and perturbed-query separation (Figure~\ref{fig:motivation}(b)). Factual reconstruction aligns the executed-query prediction with observed DINOv3 targets. One-sided separation uses this prediction as a detached reference and encourages a distinct prediction under the perturbed query. Both objectives update the shared planner, coupling observed-future supervision with trajectory-query sensitivity.

On all 220 Bench2Drive routes~\citep{jia2024bench2drive}, \method{} achieves 91.53 Driving Score and 75.45\% Success Rate. It improves on schedule-matched planning-only continuation by 5.31 DS points and nine successful routes. Matched two-seed controls on 120 routes show gains from trajectory conditioning and separation. Correct trajectory pairing improves DS by 4.25 over shuffled pairing. Prediction diagnostics on newly collected episodes connect these controls to reconstruction quality and executed-query benefits.

Our contributions are:
\begin{enumerate}
\item \textbf{Trajectory-conditioned auxiliary training for vision--language driving.} An external waypoint query conditions shared future-query states, coupling BEV supervision with a policy that directly generates driving waypoints.
\item \textbf{Joint learning from factual and unlabeled trajectory queries.} Factual reconstruction anchors the observed outcome, and one-sided separation trains the readout to respond to changes in the trajectory query.
\item \textbf{Closed-loop gains supported by matched training controls.} Full-set evaluation establishes planning performance; two-seed controls quantify the gains from trajectory conditioning, separation, and correct query pairing. Joint planning--prediction diagnostics relate these gains to reconstruction quality and query responses.
\end{enumerate}

\section{Related Work}
	\label{sec:related}

	\paragraph{World action models and video-generative priors.}
	DreamZero~\citep{ye2026dreamzero} jointly models video and robot actions with a pretrained video-diffusion backbone. Drive-WM~\citep{wang2024drivewm} synthesizes maneuver-conditioned multi-view driving videos for trajectory selection. GenCeption~\citep{wang2026genception} studies transferable visual representations and vision--language alignment from video-generative pretraining.

	\paragraph{Representation-space futures in driving.}
	LAW~\citep{li2025law} predicts perspective and BEV latents conditioned on ego trajectories. WoTE~\citep{li2025wote} and World4Drive~\citep{zheng2025world4drive} use candidate-dependent futures for online trajectory selection. UniDrive-WM~\citep{xiong2026unidrivewm}, DriveWorld-VLA~\citep{jia2026driveworldvla}, VLA-World~\citep{wang2026vlaworld}, and Latent-CoT-Drive~\citep{tan2026latentcotdrive} integrate future generation or latent imagination with VLM/VLA planning. DriveVLA-W0~\citep{li2025drivevlaw0} makes future-image prediction optional at inference, while HyWorldVLA~\citep{yu2026hyworldvla} combines latent prediction with pixel reconstruction.

	\paragraph{Planner-native semantic futures.}
	DeepSight~\citep{zhang2026deepsight} predicts DINOv3-aligned BEV futures alongside VLM reasoning and planning. Latent-WAM~\citep{wang2026latentwam} predicts latent transitions from scene and ego-state history, including commands, velocity, and acceleration, using EMA-encoder targets and omitting its dynamics module at inference. Building on DeepSight-style future supervision, \method{} introduces an external future-waypoint query into the VLM's future states and jointly trains factual reconstruction and perturbed-query separation. Its training interface combines fixed DINOv3 targets for the executed query with a separation objective for unlabeled queries.

	\paragraph{Training-time supervision and candidate evaluation.}
	EgoWAM~\citep{li2026egowam} compares pixel-, DINO-, and motion-targeted supervision. Fast-WAM~\citep{yuan2026fastwam} and SimWAM~\citep{zhao2026simwam} bypass future prediction at inference, whereas Faster-WAM~\citep{zhao2026fasterwam} retains sparse future conditioning. DA-WAM~\citep{zhong2026dawam} addresses single-outcome driving logs by supervising the expert-matched candidate with future latents and all candidates with factor, utility, and ranking targets for online evaluation. \method{} uses perturbed-query separation to train shared policy representations and deploys direct waypoint generation.

	\section{Method}
	\label{sec:method}

	\begin{figure*}[t]
		\centering
			\includegraphics[width=\textwidth]{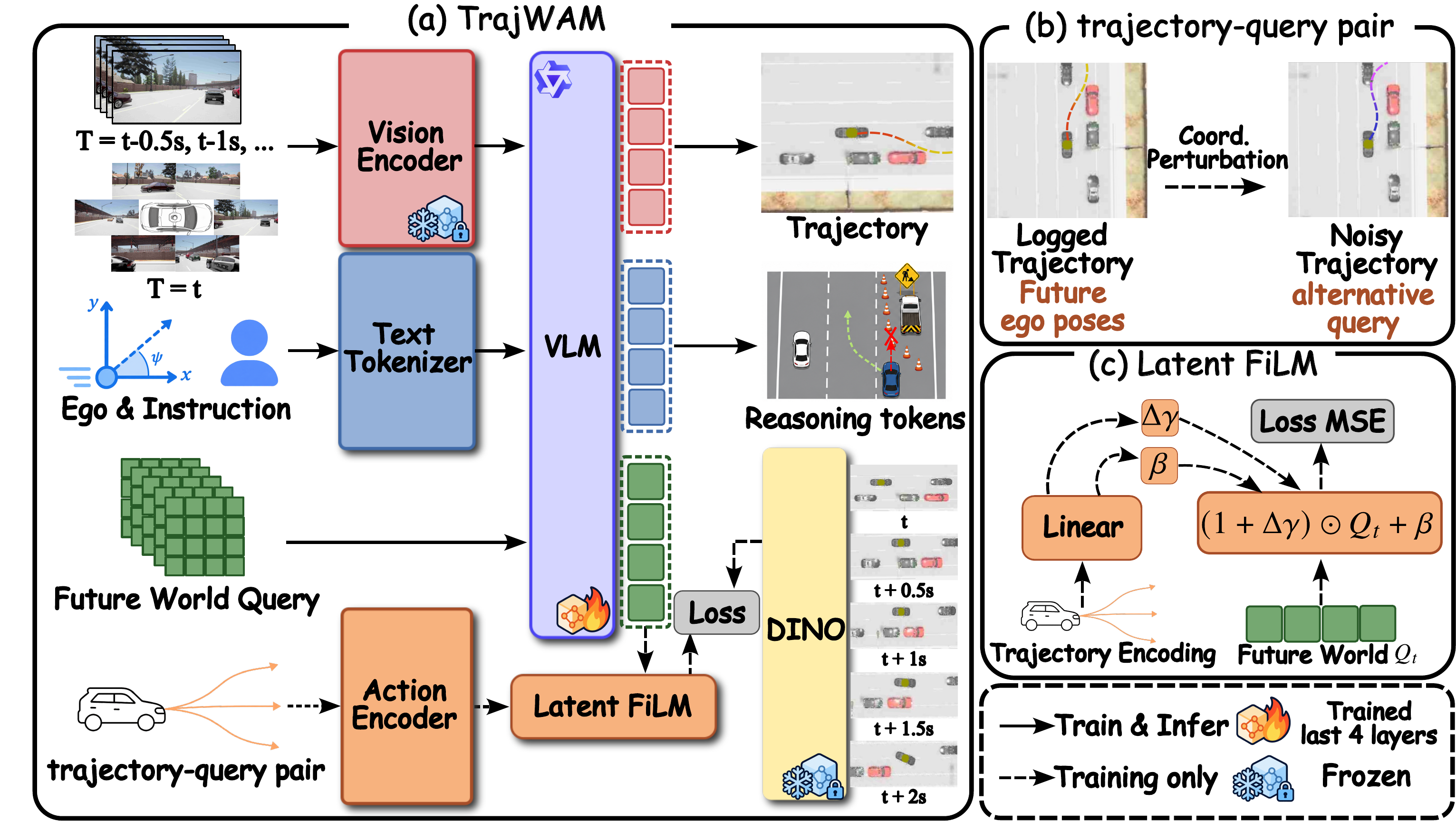}
		\caption{Overview of \method{}. (a) Shared planning and trajectory-conditioned auxiliary learning. (b) Logged and perturbed trajectory queries. (c) FiLM conditioning. Sections~\ref{sec:singleoutcome}--\ref{sec:gradient} detail the readout and objectives. Solid arrows denote training and inference; dashed arrows denote training-only computation.}
		\label{fig:overview}
	\end{figure*}

	\subsection{Problem Setup and Base Planner}

	\method{} trains an action-generating VLM with a trajectory-conditioned world readout. The readout and waypoint generator share planner states. Factual future supervision and trajectory-query separation update the trainable planner layers through this shared representation.

	At time $t$, the planner receives $X_t=(O_t,s_t,r_t,q^{\mathrm{fw}})$, where $O_t$ contains four historical front-camera frames and a current six-camera surround-view observation, $s_t$ is the ego state, $r_t$ is the route/target instruction, and $q^{\mathrm{fw}}$ is a fixed block of future-world query tokens. These tokens remain in the VLM sequence during training and inference. Their trajectory-conditioned readout operates during training. A Qwen2.5-VL-3B planner $P_\phi$ produces hidden states
	\begin{equation}
		H_t=P_\phi(X_t)\in\mathbb{R}^{L\times D},\qquad D=2048,
		\label{eq:planner_states}
	\end{equation}
	and autoregressively generates four future waypoints $\widehat A_t=\{(\widehat a_{k,1},\widehat a_{k,2})\}_{k=1}^{4}$. The inherited PID controller converts these waypoints into low-level control.

	We first train the \baseplanner{} with token-level reasoning-and-waypoint supervision and a trajectory-agnostic DINOv3-aligned BEV future-representation objective inspired by DeepSight~\citep{zhang2026deepsight}. Starting from the \baseplanner{} checkpoint, \method{} introduces the trajectory address and jointly optimizes native waypoint generation with trajectory-conditioned DINOv3-aligned BEV future-representation learning. Let
	\begin{equation}
		\mathcal T=\{0,0.5,1.0,1.5,2.0\}\ \mathrm{s},
		\label{eq:horizons}
	\end{equation}
	denote the five supervised BEV times, comprising the current anchor and four future observations aligned to the current ego frame. The aligned images $\{B_{t+\tau}^{\star}\}_{\tau\in\mathcal T}$ supply frozen DINOv3 targets for auxiliary supervision.

	In Figure~\ref{fig:overview}(a), green tokens supply future-query states to Latent FiLM, and the ActionEncoder supplies the trajectory embedding. The DINO block encodes observed BEV targets.

	\subsection{Logged and Noisy Trajectory Queries}
	\label{sec:trajectory_pair}

	Each training record supplies one realized ego trajectory
	$A_t^\star=[a_{t,1}^\star,\ldots,a_{t,4}^\star]\in\mathbb{R}^{4\times2}$,
	expressed in metres in the current ego frame, with coordinates ordered as [forward, lateral]. It supplies the factual trajectory query for auxiliary training. We construct a noisy alternative
	\begin{equation}
		\begin{aligned}
			(A_t^-)_{k,1}&=\rho_k(A_t^\star)_{k,1},
			&\rho_k&\sim\mathcal U(0.7,1.3),\\
			(A_t^-)_{k,2}&=(A_t^\star)_{k,2}+\epsilon_k,
			&\epsilon_k&\sim\mathcal N(0,2^2),
		\end{aligned}
		\qquad k\in\{1,\ldots,4\}.
		\label{eq:noisy_query}
	\end{equation}
	Forward distances are independently scaled, and lateral coordinates receive Gaussian noise. The pair $\{A_t^\star,A_t^-\}$ presents the same scene under two trajectory queries. The logged query receives observed-future supervision; the perturbed query supplies the feature-separation objective.

\subsection{Trajectory Address and Latent FiLM}
	\label{sec:address}

	For either query $A\in\{A_t^\star,A_t^-\}$, the ActionEncoder flattens and normalizes its coordinates and maps them to the VLM hidden dimension:
	\begin{equation}
		\begin{aligned}
			z_t(A)&=\operatorname{vec}(A)/\sigma,\qquad \sigma=10,\\
			e_t(A)&=W_2\operatorname{GELU}(W_1z_t(A)+b_1)+b_2
			\in\mathbb{R}^{D}.
		\end{aligned}
		\label{eq:action_encoder}
	\end{equation}
	Let $M_{t,\tau}^{\mathrm{fw}}$ select the valid future-world query tokens at horizon $\tau$ and let $Q_{t,\tau}=H_t[M_{t,\tau}^{\mathrm{fw}}]$. The same sample-level address is broadcast over all valid tokens and all five horizons. Latent FiLM produces
	\begin{equation}
		\begin{aligned}
			[\Delta\gamma_t(A),\beta_t(A)]&=F_{\mathrm{lat}}(e_t(A)),\\
			\widetilde Q_{t,\tau}(A)&=
			\left(1+\Delta\gamma_t(A)\right)\odot Q_{t,\tau}+\beta_t(A).
		\end{aligned}
		\label{eq:latent_film}
	\end{equation}
	Zero-initializing the FiLM projection gives the identity transform. Trajectory conditioning is applied to the auxiliary readout after planner-logit computation.

	\subsection{DINOv3-Aligned Prediction from Single-Outcome Logs}
	\label{sec:singleoutcome}

	The readout between Latent FiLM and Loss in Figure~\ref{fig:overview}(a) uses a DINO-space projection $V:\mathbb{R}^{D}\rightarrow\mathbb{R}^{1024}$, learned with the \baseplanner{} and frozen during \method{} training:
	\begin{equation}
		\begin{aligned}
			\widehat Y_{t,\tau}(A)
			&=V\!\left(\widetilde Q_{t,\tau}(A)\right),\\
			\widehat{\mathbf Y}_t(A)
			&=\{\widehat Y_{t,\tau}(A)\}_{\tau\in\mathcal T}.
		\end{aligned}
		\label{eq:semantic_prediction}
	\end{equation}
	A frozen DINOv3 ViT-L/16 teacher constructs the factual targets
	\begin{equation}
		Y_{t,\tau}^\star=
		\sg\!\left[\operatorname{DINOv3}\!\left(B_{t+\tau}^\star\right)\right]_{\Omega},
		\qquad \tau\in\mathcal T,
		\label{eq:dino_target}
	\end{equation}
	where $\Omega$ selects the CLS token and valid patch tokens; register tokens are excluded. ``Loss MSE'' in Figure~\ref{fig:overview}(c) is computed after projection:
	\begin{equation}
		\mathcal L_{\mathrm{fact}}=
		\operatorname{MSE}_{\tau,\Omega}\!\left(
		\widehat Y_{t,\tau}(A_t^\star),Y_{t,\tau}^\star\right).
		\label{eq:factual}
	\end{equation}
	The unlabeled query $A_t^-$ is trained through the valid-token feature distance
	\begin{equation}
			d_{\mathrm{sem}}(U,Z)=
			\operatorname{mean}_{\tau,\Omega,c}|U_{\tau,\Omega,c}-Z_{\tau,\Omega,c}|,
		\label{eq:semantic_distance}
	\end{equation}
	where $c\in\{1,\ldots,1024\}$ indexes the raw, unnormalized DINO feature channels. We then use the one-sided separation objective
	\begin{equation}
		\mathcal L_{\mathrm{sep}}=
		\left[m_{\mathrm{sem}}-
		d_{\mathrm{sem}}\!\left(
		\sg\!\left(\widehat{\mathbf Y}_t(A_t^\star)\right),
			\widehat{\mathbf Y}_t(A_t^-)\right)\right]_+,
		\label{eq:lsep}
	\end{equation}
	where $[x]_+=\max(x,0)$. The factual prediction anchors the separation margin under stop-gradient; gradients flow through the perturbed prediction.

	\subsection{Objective, Gradient Routing, and Native Inference}
	\label{sec:gradient}

	The auxiliary ``Loss'' in Figure~\ref{fig:overview}(a) combines factual reconstruction and separation; the full objective includes planning supervision:
	\begin{equation}
		\mathcal L=\mathcal L_{\mathrm{rec}}+2\mathcal L_{\mathrm{fact}}
		+\lambda_{\mathrm{sep}}\mathcal L_{\mathrm{sep}},
		\label{eq:total}
	\end{equation}
	where $\mathcal L_{\mathrm{rec}}$ is token-level cross-entropy for reasoning and waypoint generation, $m_{\mathrm{sem}}$ is the DINO-space margin, and $\lambda_{\mathrm{sep}}$ weights separation. Both auxiliary losses update the ActionEncoder, latent FiLM, and final four trainable planner layers. Factual reconstruction differentiates through the logged-query prediction; separation differentiates through the perturbed-query prediction and uses the logged prediction as a detached reference. The frozen DINO projection transmits gradients to its input states. The vision tower and multimodal projector remain frozen, and the frozen DINOv3 teacher supplies targets under stop-gradient. The planning loss updates the trainable planner layers.

Native inference runs autoregressive reasoning and waypoint generation followed by the inherited PID controller. The fixed future-query tokens remain in the prompt, and the auxiliary trajectory-conditioned readout is bypassed. On an RTX 5090, median generation takes 3.5086\,s for \method{} and 3.4926\,s for public DeepSight (Appendix~\ref{app:native_profile}).

\begin{table*}[t]
		\centering
		\caption{Bench2Drive Base Set closed-loop results. Venue reports conference/preprint and year; $^{*}$ denotes expert-feature distillation. Bold/underlined values mark the best/second-best listed results.}
		\label{tab:bench2drive_base}

		\renewcommand{\arraystretch}{1.08}
		\setlength{\tabcolsep}{3.5pt}

		\resizebox{\textwidth}{!}{
			\begin{tabular}{@{}lccc*{4}{c}@{}}
				\toprule

				\multirow{2}{*}{Method}
				&
				\multirow{2}{*}{Venue}
				&
				\multirow{2}{*}{Paradigm}
				&
				\multirow{2}{*}{Expert}
				&
				\multicolumn{4}{c}{Closed-loop}
				\\

				\cmidrule(lr){5-8}

				& & & &
				\cellcolor{metricgray}DS$\uparrow$
				&
				\cellcolor{metricgray}SR(\%)$\uparrow$
				&
				Efficiency$\uparrow$
				&
				Smoothness$\uparrow$
				\\

				\midrule

				TCP*~\citep{wu2022tcp} & NeurIPS~'22
				& E2E
				& Think2Drive
				& \cellcolor{metricgray}40.70
				& \cellcolor{metricgray}15.00
				& 54.26
				& \underline{47.80}
				\\

				ThinkTwice*~\citep{jia2023thinktwice} & CVPR~'23
				& E2E
				& Think2Drive
				& \cellcolor{metricgray}62.44
				& \cellcolor{metricgray}31.23
				& 69.33
				& 16.22
				\\

				DriveAdapter*~\citep{jia2023driveadapter} & ICCV~'23
				& E2E
				& Think2Drive
				& \cellcolor{metricgray}64.22
				& \cellcolor{metricgray}33.08
				& 70.22
				& 16.01
				\\

				VAD~\citep{jiang2023vad} & ICCV~'23
				& E2E
				& Think2Drive
				& \cellcolor{metricgray}42.35
				& \cellcolor{metricgray}15.00
				& 157.94
				& 46.01
				\\

				GenAD~\citep{zheng2024genad} & ECCV~'24
				& E2E
				& Think2Drive
				& \cellcolor{metricgray}44.81
				& \cellcolor{metricgray}15.90
				& -
				& -
				\\

				MomAD~\citep{song2025don} & CVPR~'25
				& E2E
				& Think2Drive
				& \cellcolor{metricgray}44.54
				& \cellcolor{metricgray}16.71
				& 170.21
				& \textbf{48.63}
				\\

				DriveTransformer~\citep{jia2025drivetransformer} & ICLR~'25
				& E2E
				& Think2Drive
				& \cellcolor{metricgray}63.46
				& \cellcolor{metricgray}35.01
				& 100.64
				& 20.78
				\\

				DiffusionDrive~\citep{liao2025diffusiondrive} & CVPR~'25
				& E2E
				& PDM-Lite
				& \cellcolor{metricgray}77.68
				& \cellcolor{metricgray}52.72
				& -
				& -
				\\

				SparseDriveV2~\citep{sun2026sparsedrivev2} & ECCV~'26
				& E2E
				& Think2Drive
				& \cellcolor{metricgray}\underline{89.15}
				& \cellcolor{metricgray}70.00
				& 199.84
				& 18.32
				\\

				\midrule

				ReasonPlan~\citep{liu2025reasonplan} & CoRL~'25
				& VLM
				& Think2Drive
				& \cellcolor{metricgray}64.01
				& \cellcolor{metricgray}34.55
				& 180.64
				& 25.63
				\\

				ORION~\citep{fu2025orion} & ICCV~'25
				& VLM
				& Think2Drive
				& \cellcolor{metricgray}77.74
				& \cellcolor{metricgray}54.62
				& 151.48
				& 17.38
				\\

				SimLingo~\citep{renz2025simlingo} & CVPR~'25
				& VLM
				& PDM-Lite
				& \cellcolor{metricgray}85.94
				& \cellcolor{metricgray}66.82
				& \textbf{244.18}
				& 25.49
				\\

				AutoVLA~\citep{zhou2026autovla} & NeurIPS~'25
				& VLM
				& Think2Drive
				& \cellcolor{metricgray}78.84
				& \cellcolor{metricgray}57.73
				& 146.93
				& 39.33
				\\

				DeepSight~\citep{zhang2026deepsight} & ICML~'26
				& VLM
				& Think2Drive
				& \cellcolor{metricgray}86.23
				& \cellcolor{metricgray}71.36
				& \underline{201.71}
				& 16.11
				\\

				\midrule

				UniDrive-WM~\citep{xiong2026unidrivewm} & ECCV~'26 & WM & Think2Drive & \cellcolor{metricgray}79.31 & \cellcolor{metricgray}56.42 & 158.65 & 27.93 \\

				GeRo-Qwen2.5VL~\citep{yasarla2026generative} & CVPR~'26 & WM & Think2Drive & \cellcolor{metricgray}79.60 & \cellcolor{metricgray}57.80 & 159.66 & 44.75 \\

				DynVLA~\citep{shang2026dynvla} & ICML~'26 & WM & Think2Drive & \cellcolor{metricgray}88.34 & \cellcolor{metricgray}\underline{72.73} & - & - \\

				LVDrive~\citep{mei2026lvdrive} & arXiv~'26 & WAM & Think2Drive & \cellcolor{metricgray}80.71 & \cellcolor{metricgray}58.26 & 155.77 & 14.34 \\

				\rowcolor{oursblue}
				\fullmodel\:(Ours) & This work
				& WAM
				& Think2Drive
				& \textbf{91.53}
				& \textbf{75.45}
				& 171.41
				& 15.27
				\\

				\bottomrule
			\end{tabular}
		}
	\end{table*}

\begin{table*}[t]
		\centering
		\caption{Bench2Drive multi-ability results (\%). Venue and expert-feature distillation ($^{*}$) follow Table~\ref{tab:bench2drive_base}. Bold/underlined values mark the best/second-best listed results.}
		\label{tab:bench2drive_multi_ability}

		\renewcommand{\arraystretch}{1.08}
		\setlength{\tabcolsep}{3.5pt}

		\resizebox{\textwidth}{!}{
			\begin{tabular}{@{}lcc*{6}{c}@{}}
				\toprule

				\multirow{2}{*}{Method}
				&
				\multirow{2}{*}{Venue}
				&
				\multirow{2}{*}{Paradigm}
				&
				\multicolumn{6}{c}{Ability (\%) $\uparrow$}
				\\

				\cmidrule(lr){4-9}

				& & &
				Merging
				&
				Overtaking
				&
				Emergency Brake
				&
				Give Way
				&
				Traffic Sign
				&
				\cellcolor{metricgray}Mean
				\\

				\midrule

				TCP*~\citep{wu2022tcp} & NeurIPS~'22
				& E2E
				& 16.18
				& 20.00
				& 20.00
				& 10.00
				& 6.99
				& \cellcolor{metricgray}14.63
				\\

				ThinkTwice*~\citep{jia2023thinktwice} & CVPR~'23
				& E2E
				& 27.38
				& 18.42
				& 35.82
				& \textbf{50.00}
				& 54.23
				& \cellcolor{metricgray}37.17
				\\

				DriveAdapter*~\citep{jia2023driveadapter} & ICCV~'23
				& E2E
				& 28.82
				& 26.38
				& 48.76
				& \textbf{50.00}
				& 56.43
				& \cellcolor{metricgray}42.08
				\\

				AD-MLP~\citep{zhai2023rethinking} & arXiv~'23
				& E2E
				& 0.00
				& 0.00
				& 0.00
				& 0.00
				& 4.35
				& \cellcolor{metricgray}0.87
				\\

				UniAD-Base~\citep{hu2023planning} & CVPR~'23
				& E2E
				& 14.10
				& 17.78
				& 21.67
				& 10.00
				& 14.21
				& \cellcolor{metricgray}15.55
				\\

				VAD~\citep{jiang2023vad} & ICCV~'23
				& E2E
				& 8.11
				& 24.44
				& 18.64
				& 20.00
				& 19.15
				& \cellcolor{metricgray}18.07
				\\

				DriveTransformer~\citep{jia2025drivetransformer} & ICLR~'25
				& E2E
				& 17.57
				& 35.00
				& 48.36
				& \underline{40.00}
				& 52.10
				& \cellcolor{metricgray}38.60
				\\

				SparseDriveV2~\citep{sun2026sparsedrivev2} & ECCV~'26
				& E2E
				& \underline{66.25}
				& 75.55
				& 75.00
				& \textbf{50.00}
				& 71.57
				& \cellcolor{metricgray}67.67
				\\

				\midrule

				ReasonPlan~\citep{liu2025reasonplan} & CoRL~'25
				& VLM
				& 37.50
				& 26.67
				& 33.30
				& \underline{40.00}
				& 45.79
				& \cellcolor{metricgray}36.66
				\\

				ORION~\citep{fu2025orion} & ICCV~'25
				& VLM
				& 25.00
				& 71.11
				& 78.33
				& 30.00
				& 69.15
				& \cellcolor{metricgray}54.72
				\\

				SimLingo~\citep{renz2025simlingo} & CVPR~'25
				& VLM
				& 60.00
				& 60.00
				& 78.33
				& \textbf{50.00}
				& \textbf{77.89}
				& \cellcolor{metricgray}65.25
				\\

				DeepSight~\citep{zhang2026deepsight} & ICML~'26
				& VLM
				& 60.00
				& \textbf{91.11}
				& 78.33
				& \textbf{50.00}
				& 71.58
				& \textbf{70.20}
				\\

				\midrule

				UniDrive-WM~\citep{xiong2026unidrivewm} & ECCV~'26
				& WM
				& 29.97
				& 74.64
				& \underline{79.98}
				& \underline{40.00}
				& 71.54
				& 59.23
				\\

				GeRo-Qwen2.5VL~\citep{yasarla2026generative} & CVPR~'26
				& WM
				& 37.83
				& 72.81
				& \textbf{83.10}
				& \textbf{50.00}
				& 66.17
				& 61.98
				\\

				LVDrive~\citep{mei2026lvdrive} & arXiv~'26
				& WAM
				& 39.74
				& 69.89
				& 71.67
				& 20.00
				& 74.21
				& 54.90
				\\

				\rowcolor{oursblue}
				\fullmodel\:(Ours) & This work
				& WAM
				& \textbf{75.00}
				& \underline{82.22}
				& 65.00
				& \textbf{50.00}
				& \underline{77.37}
				& \underline{69.92}
				\\

				\bottomrule
			\end{tabular}
		}
	\end{table*}

\begin{table*}[t]
		\centering
		\setlength{\tabcolsep}{3pt}
		\caption{Phase-II component ablation on 220 routes under the setup in Section~\ref{sec:experiments}. Factual DINO denotes target reconstruction, unconditioned in \futureft{}; Traj.-FiLM comprises the ActionEncoder and latent FiLM; Sep. denotes Equation~\ref{eq:lsep}. The \planningft{} control is shared with Table~\ref{tab:matched_control}.}
		\label{tab:component_ablation}
		\begin{tabular}{@{}l|ccc|cc@{}}
			\toprule
			Model & \shortstack{Factual DINO} & \shortstack{Traj.-FiLM} & \shortstack{Sep.} & DS$\uparrow$ & SR$\uparrow$ \\
			\midrule
			\shortstack[l]{\planningft{}} & -- & -- & -- & 86.22 & 71.36\% \\
			\shortstack[l]{\futureft{}} & \checkmark & -- & -- & 85.34 & 73.18\% \\
			\nosepmodel{} & \checkmark & \checkmark & -- & 89.34 & 74.55\% \\
			\rowcolor{oursblue}
			\fullmodel{} & \checkmark & \checkmark & \checkmark & \textbf{91.53} & \textbf{75.45\%} \\
			\bottomrule
		\end{tabular}
	\end{table*}

\begin{table*}[t]
\centering
\caption{Matched two-seed DS comparisons on the fixed 120-route subset. Summary values are mean $\pm$ sample standard deviation across seeds 42 and 43. All rows share the Phase-I initialization and 24-epoch training budget.}
\label{tab:subset120_summary}
\begin{tabular}{@{}lccc@{}}
\toprule
Training variant & Seed 42 DS $\uparrow$ & Seed 43 DS $\uparrow$ & Mean DS $\pm$ SD \\
\midrule
Planning-only FT & 82.64 & 83.16 & $82.90\pm0.37$ \\
Unconditioned future FT & 82.11 & 82.49 & $82.30\pm0.27$ \\
\method{} w/o separation & 86.28 & 86.52 & $86.40\pm0.17$ \\
\method{} (additive) & 86.07 & 85.83 & $85.95\pm0.17$ \\
\rowcolor{oursblue} \method{} (correct pairing) & \textbf{88.69} & \textbf{88.21} & $\mathbf{88.45\pm0.34}$ \\
Shuffled pairing & 84.02 & 84.38 & $84.20\pm0.25$ \\
\bottomrule
\end{tabular}
\end{table*}

\section{Results}
	\label{sec:results}

	\subsection{Experimental Setup}
	\label{sec:experiments}

	Phase I trains our in-house \baseplanner{} from Qwen2.5-VL-3B using autoregressive reasoning/waypoint supervision and trajectory-agnostic DINOv3 future reconstruction, following the DeepSight-style recipe. Phase II adds trajectory conditioning and perturbed-query supervision to this shared starting point.

	Phase II starts from this checkpoint and trains \fullmodel{} for 24 epochs on 226,656 Bench2Drive examples. We use eight A100 GPUs, BF16, an effective batch size of 64, AdamW with cosine decay and 10\% warm-up, and training seed 42 for the 220-route evaluation. Learning rates are $2\times10^{-5}$ for the last four planner layers and $2\times10^{-4}$ for \fullmodel{} modules; no validation-based checkpoint selection is used. Optimization details appear in Appendix~\ref{app:implementation}; checkpoint parameter counts appear in Appendix~\ref{app:footprint}.

	\paragraph{Model variants.}
	Our in-house \baseplanner{} initializes all four Phase-II variants in Tables~\ref{tab:matched_control} and~\ref{tab:component_ablation}; each is independently fine-tuned (FT) for 24 epochs. \planningft{} optimizes reasoning and waypoint generation. \futureft{} adds trajectory-agnostic factual DINOv3 reconstruction. \nosepmodel{} conditions factual reconstruction on the trajectory query, and \fullmodel{} adds the separation objective. All four variants share the Phase-I DINOv3 training history. Table~\ref{tab:bench2drive_base} compares them with published DeepSight results and other driving methods.

	Closed-loop evaluation covers all 220 routes in the Bench2Drive Base Set and reports the official Driving Score (DS) and strict Success Rate (SR). DS averages route completion weighted by infraction penalties, and SR counts strictly successful routes. For the 220-route results in Tables~\ref{tab:bench2drive_base}, \ref{tab:bench2drive_multi_ability}, \ref{tab:component_ablation}, and~\ref{tab:matched_control}, each Phase-II variant uses seed 42 and one complete simulator sweep. Table~\ref{tab:subset120_summary} uses seeds 42 and 43, with one sweep of the same fixed 120 routes per checkpoint. Prior-method scores come from their publications. Appendix~\ref{app:dino_metrics} specifies the checkpoints, new episodes, and protocol for prediction diagnostics.

	\subsection{Closed-Loop Planning and Phase-II Controls}
	\label{sec:matched_control}

	\paragraph{Overall closed-loop performance.}
	\fullmodel{} achieves the highest DS (91.53) and SR (75.45\%) among the methods in Table~\ref{tab:bench2drive_base}. It exceeds the next-best DS of 89.15 from SparseDriveV2 by 2.38 points and the next-best SR of 72.73\% from DynVLA by 2.72 percentage points. \fullmodel{} also obtains 171.41 Efficiency and 15.27 Smoothness.

    \paragraph{Multi-ability performance.}
	\fullmodel{} leads Merging (75.00\%) and ties the highest Give Way score (50.00\%) in Table~\ref{tab:bench2drive_multi_ability}. It ranks second in Overtaking (82.22\%) and Traffic Sign (77.37\%). Across the five abilities, its mean is 69.92\%, second overall and 0.28 percentage points below DeepSight. Emergency Brake scores are 65.00\% for \fullmodel{}, 78.33\% for DeepSight, and 83.10\% for GeRo-Qwen2.5VL.

	\fullmodel{} improves over the schedule-matched \planningft{} control by $5.31$ DS points and $4.09$ percentage points in SR, adding nine successful routes (Appendix Table~\ref{tab:matched_control}). Table~\ref{tab:component_ablation} compares the contributions of factual reconstruction, trajectory conditioning, and separation under the shared Phase-II setup.

	\paragraph{Component ablation.}
	The trajectory-agnostic \futureft{} reference reaches 85.34 DS and 73.18\% SR (161/220), compared with 86.22 DS and 71.36\% SR (157/220) for \planningft{}.

	Adding trajectory conditioning yields \nosepmodel{} with 89.34 DS and 74.55\% SR (164/220): a gain of 4.00 DS points and three successful routes over \futureft{}. Adding separation further raises performance to 91.53 DS and 75.45\% SR (166/220), contributing another 2.19 DS points and two successful routes.

\paragraph{Multi-seed subset evaluation.}
The planning gains hold across both training seeds on a fixed 120-route subset covering all 44 scenario types (Table~\ref{tab:subset120_summary}). \method{} achieves $88.45\pm0.34$ DS and $68.75\pm0.59\%$ SR across seeds 42 and 43. Mean gains are 5.55 DS over planning-only FT, 2.05 DS over \nosepmodel{}, and 2.50 DS over matched additive conditioning (Appendix~\ref{app:subset120}). Across the four paired component comparisons, conditional 95\% intervals lie above zero for DS and include zero for SR (Appendix Table~\ref{tab:subset120_paired}).

\subsection{Future-Representation and Trajectory-Query Diagnostics}
\label{sec:prediction_diagnostics}

We evaluate prediction quality and query correspondence using the same checkpoints as the matched planning controls. The evaluation comprises 374 windows from 15 newly collected episodes across nine maps; 371 windows have fixed, speed- and direction-matched alternative queries. All variants and both training seeds share observations, targets, and query pairs (Appendix~\ref{app:matched_prediction}).

\begin{figure}[!ht]
\centering
\includegraphics[width=\linewidth]{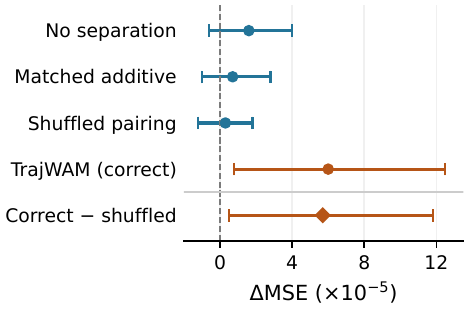}
\caption{Executed-query benefits under different training variants, averaged over seeds 42 and 43 and 15 routes. Positive $\Delta$MSE favors executed over mismatched queries on 371 fixed windows. Bars show conditional route-bootstrap 95\% intervals; the bottom row directly compares correct and shuffled training.}
\label{fig:paired_routes}
\end{figure}

Correct trajectory--scene pairing improves training: relative to within-group shuffled pairing, the complete method gains 4.25 DS across seeds 42 and 43, with a paired 95\% interval of $[2.64,5.89]$; SR increases by 1.67 percentage points (Appendix~\ref{app:training_pairing}). Its executed-query benefit is $6.00\times10^{-5}$ MSE, exceeding shuffled training by $5.70\times10^{-5}$, with a paired interval of $[0.50,11.80]\times10^{-5}$ (Figure~\ref{fig:paired_routes}).

Trajectory conditioning reduces future MSE from 0.03110 to 0.02985 and improves DS by 4.10 over unconditioned supervision. Adding separation further raises DS by 2.05, while MSE becomes 0.03020 (Table~\ref{tab:matched_prediction}). Together, these controls support trajectory-aligned auxiliary supervision for policy learning: conditioning improves both prediction and planning, and separation adds planning gains with higher aggregate MSE. Fixed-query responses vary across scenes (Appendix~\ref{app:query_precision}).

\begin{figure*}[p]
\centering
\includegraphics[width=\textwidth,height=0.88\textheight,keepaspectratio]{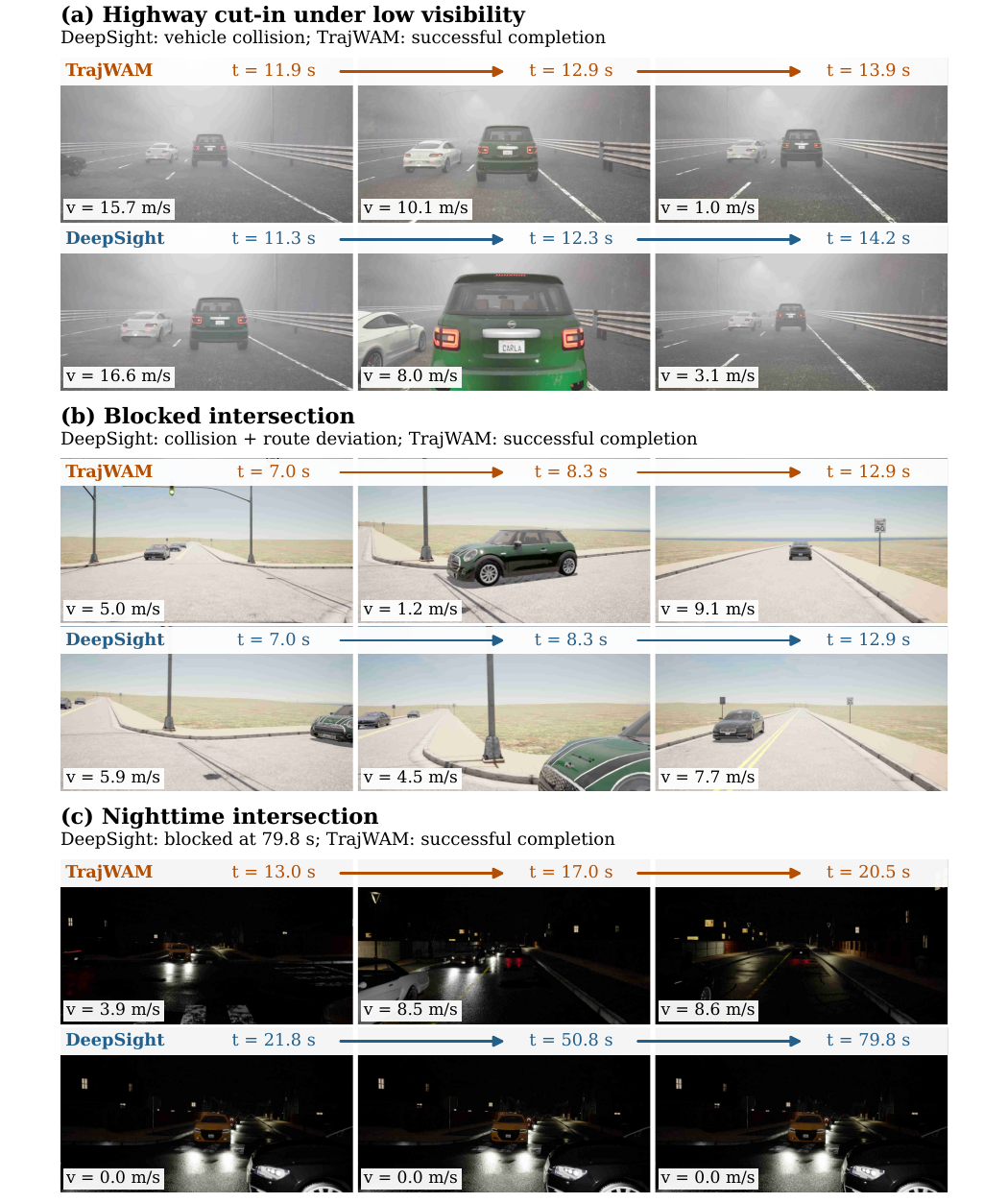}
\caption{Selected closed-loop comparisons with DeepSight under matching route, scenario, and weather configurations. Each case shows \method{} above DeepSight. Recorded DeepSight outcomes are (a) vehicle collision, (b) collision and route deviation, and (c) blocked-agent termination; \method{} completes all three routes successfully. Arrows indicate chronological order within each row, with actual simulation times and speeds. Columns use nearby ego positions in (a), equal elapsed times in (b), and independent time sequences in (c) to show progression versus persistent standstill.}
\label{fig:closed_loop_qualitative}
\end{figure*}

\paragraph{Qualitative closed-loop comparisons.}
Figure~\ref{fig:closed_loop_qualitative} compares three selected routes under matching scenario and weather configurations. \method{} slows for the highway cut-in and crossing traffic, then completes both routes; DeepSight records collisions and, in the second case, route deviation. At the nighttime intersection, \method{} proceeds through traffic while DeepSight remains stopped until blocked-agent termination.

\section{Conclusion}
\method{} turns observed futures and unlabeled trajectory queries into joint auxiliary supervision for vision--language driving. Factual reconstruction and one-sided separation train shared planner representations, and the deployed planner directly generates waypoints. On all 220 Bench2Drive routes, it achieves 91.53 DS and 75.45\% SR, improving on planning-only continuation by 5.31 DS points and nine successful routes. Matched training controls establish the contribution of trajectory conditioning, separation, and correct query pairing to these planning gains.

\clearpage
\subsection*{LLM Usage Statement}
    GPT-6 was used solely to refine author-written text and code. The authors reviewed and verified all changes and take full responsibility for the manuscript.

{\small
\bibliographystyle{ieeenat_fullname}
\bibliography{trajscope_references}
}
\clearpage
\appendix
\section{Additional Implementation Details}
	\label{app:implementation}

	\paragraph{DINO readout and query construction.}
	Equations~\ref{eq:action_encoder}--\ref{eq:semantic_prediction} specify the DINO readout. The ActionEncoder uses $\sigma=10$; FiLM is zero-initialized around the identity and broadcasts one trajectory code over all valid query tokens and five horizons. A frozen DINOv3 teacher supplies BEV supervision, using the same valid CLS-and-patch-token mask as the prediction. For $A_t^-$, the forward coordinate is independently scaled in $[0.7,1.3]$ and the lateral coordinate receives Gaussian noise with standard deviation $2\,\mathrm{m}$, as stated in Equation~\ref{eq:noisy_query}. The configuration uses $m_{\mathrm{sem}}=0.05$ and $\lambda_{\mathrm{sep}}=0.5$.

\paragraph{Gradient routing.}
	Let $\phi_{\mathrm{adapt}}$ denote the final four trainable planner layers and $\theta$ the ActionEncoder and latent-FiLM parameters. The training gradients are
	\begin{align}
		\nabla_{\phi_{\mathrm{adapt}}}\mathcal L
		&=\nabla_{\phi_{\mathrm{adapt}}}\!\left(\mathcal L_{\mathrm{rec}}+2\mathcal L_{\mathrm{fact}}+0.5\mathcal L_{\mathrm{sep}}\right),\\
		\nabla_\theta\mathcal L
		&=\nabla_\theta\!\left(2\mathcal L_{\mathrm{fact}}+0.5\mathcal L_{\mathrm{sep}}\right),
		\qquad \nabla_\theta\mathcal L_{\mathrm{rec}}=0.
	\end{align}
	The external trajectory conditions the DINO readout after planner-logit computation. Both factual reconstruction and perturbed-query separation backpropagate through the DINO projection and FiLM to $\phi_{\mathrm{adapt}}$. Equations~\ref{eq:dino_target} and~\ref{eq:lsep} identify the detached teacher targets and factual separation reference.

	\paragraph{Optimization.}
	Training uses AdamW with cosine decay, 10\% warm-up, BF16, and an effective batch size of 64 across eight A100 GPUs. The 220-route evaluation uses training seed 42; the matched 120-route comparisons use seeds 42 and 43. The final four planner layers use learning rate $2\times10^{-5}$ and the ActionEncoder and latent FiLM use $2\times10^{-4}$. These are the trainable components during adaptation.

	\section{Additional Future-World Prediction Diagnostics}
	\label{app:future_world_qualitative}

\begin{figure*}[p]
		\centering
		\includegraphics[width=\linewidth]{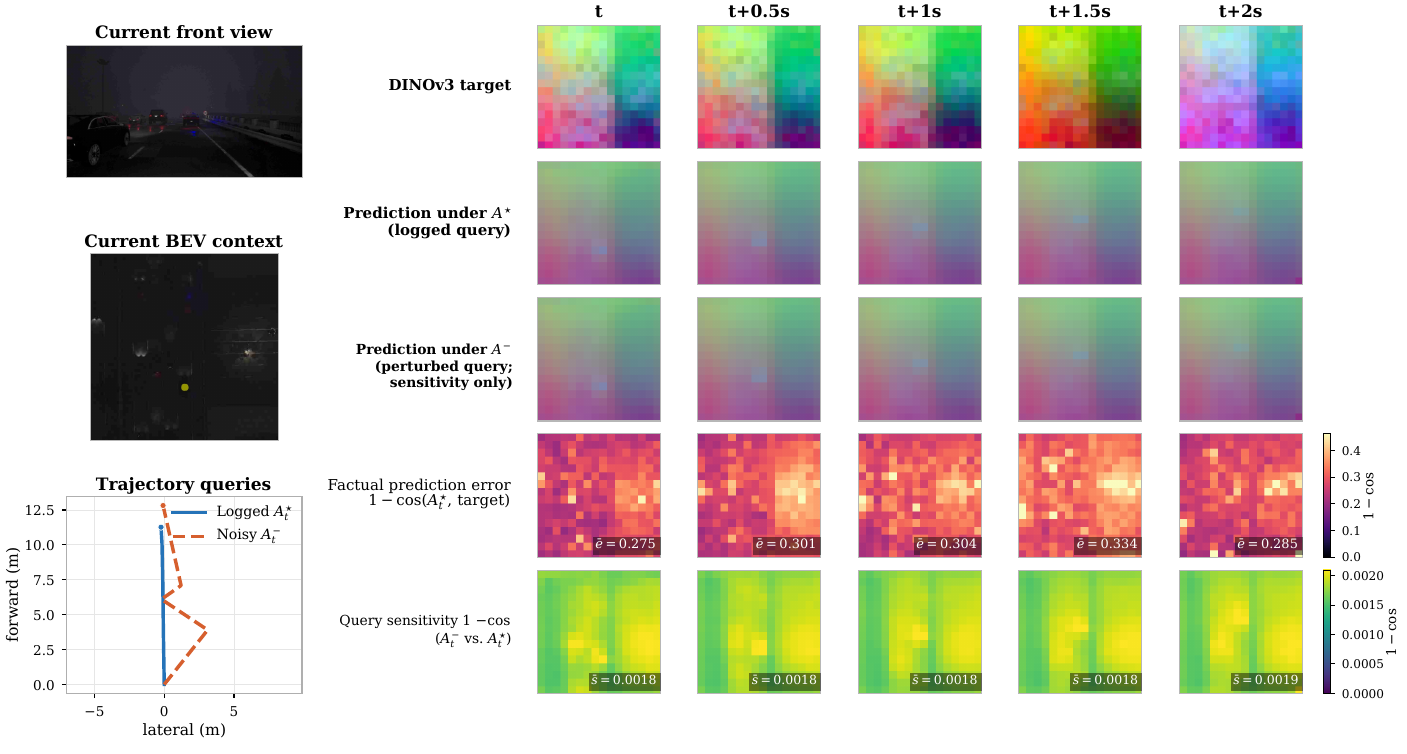}
		\vspace{0.5ex}
		\includegraphics[width=\linewidth]{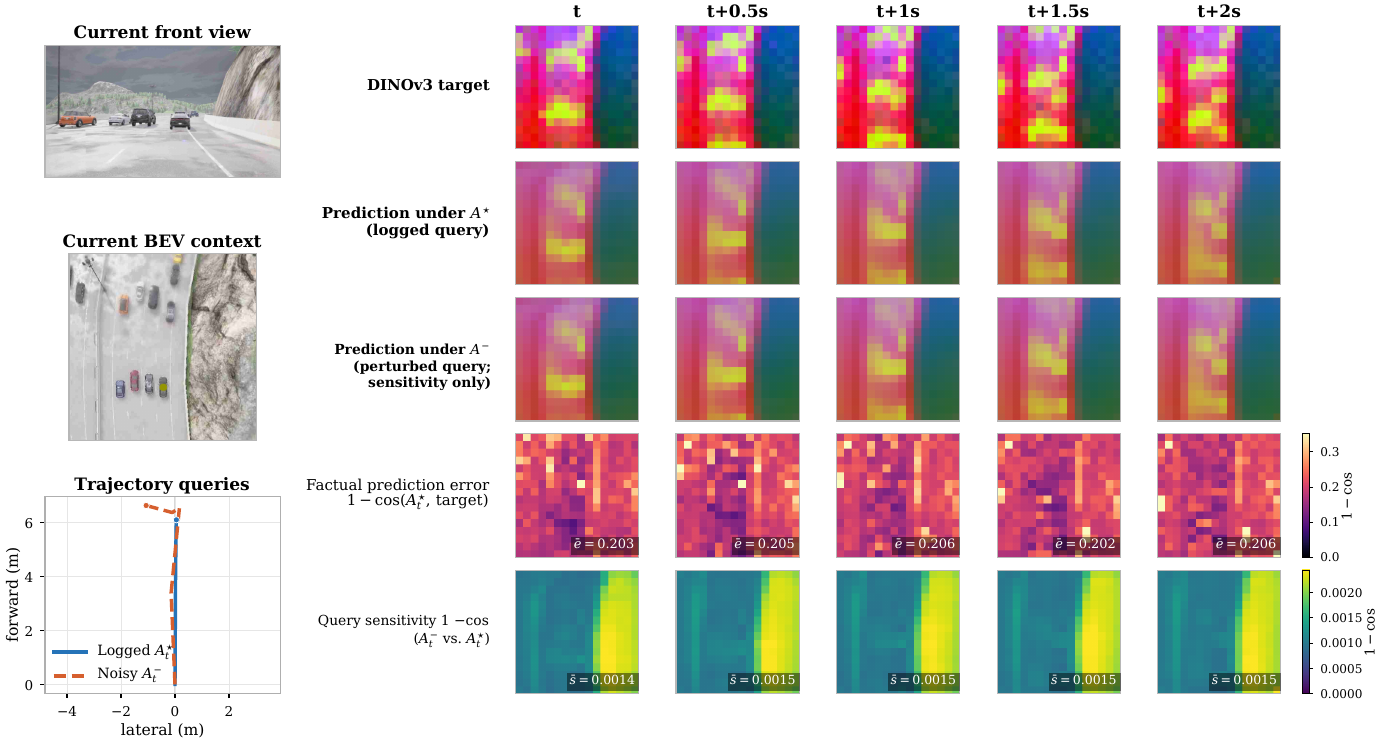}
\caption{Future-feature predictions under longitudinal deceleration perturbations. \emph{Top:} rainy example from \texttt{Accident\_Town04\_Route205\_Weather23}, frame 51, perturbation seed 2042. \emph{Bottom:} snowy example from \texttt{Accident\_Town04\_Route160\_Weather3}, frame 51, seed 2029. Columns span $t$ through $t+2\,$s in $0.5\,$s increments. Rows show the factual DINOv3 target, predictions under the logged and perturbed queries, factual cosine error, and label-free query sensitivity. Appendix~\ref{app:future_world_qualitative} describes the shared-PCA visualization and native-space metric computation.}
		\label{fig:future_world_deceleration}
	\end{figure*}

\begin{figure*}[p]
		\centering
		\includegraphics[width=\linewidth]{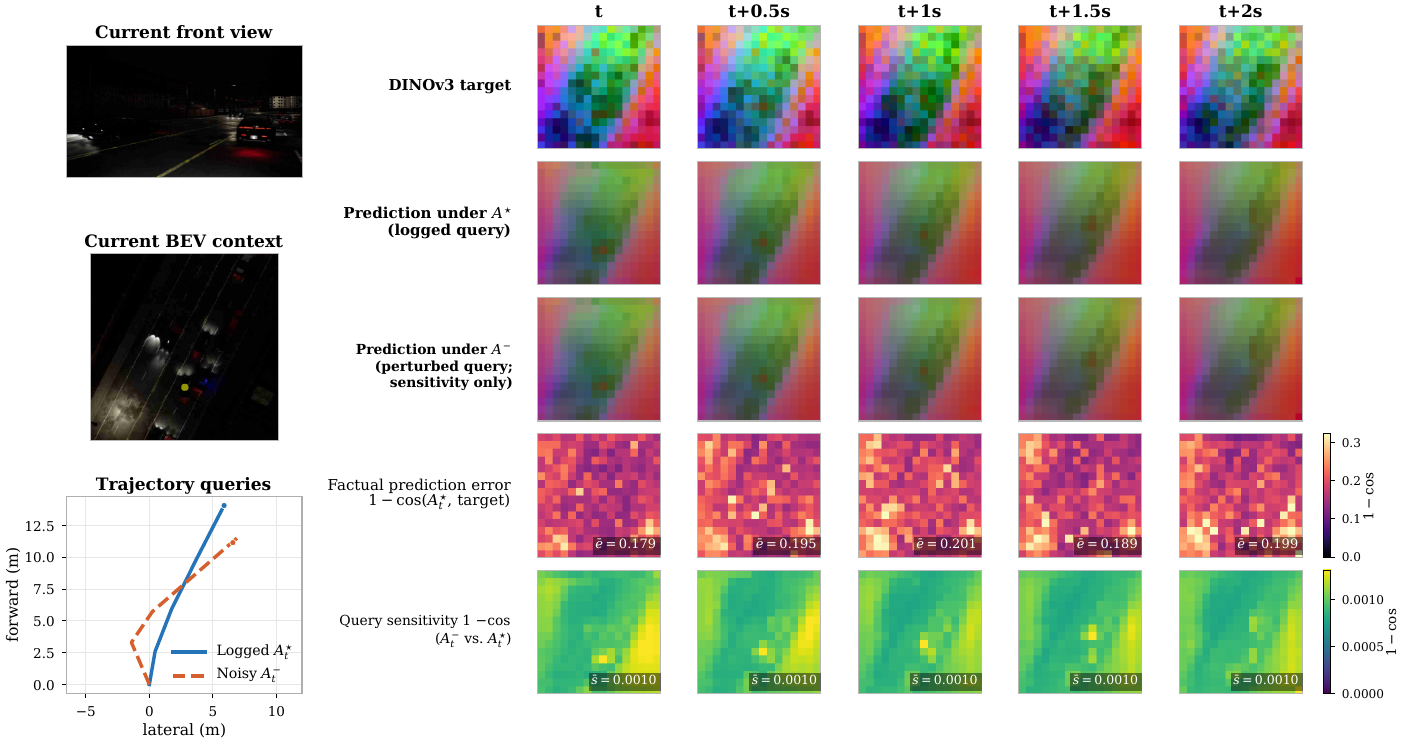}
		\vspace{0.5ex}
		\includegraphics[width=\linewidth]{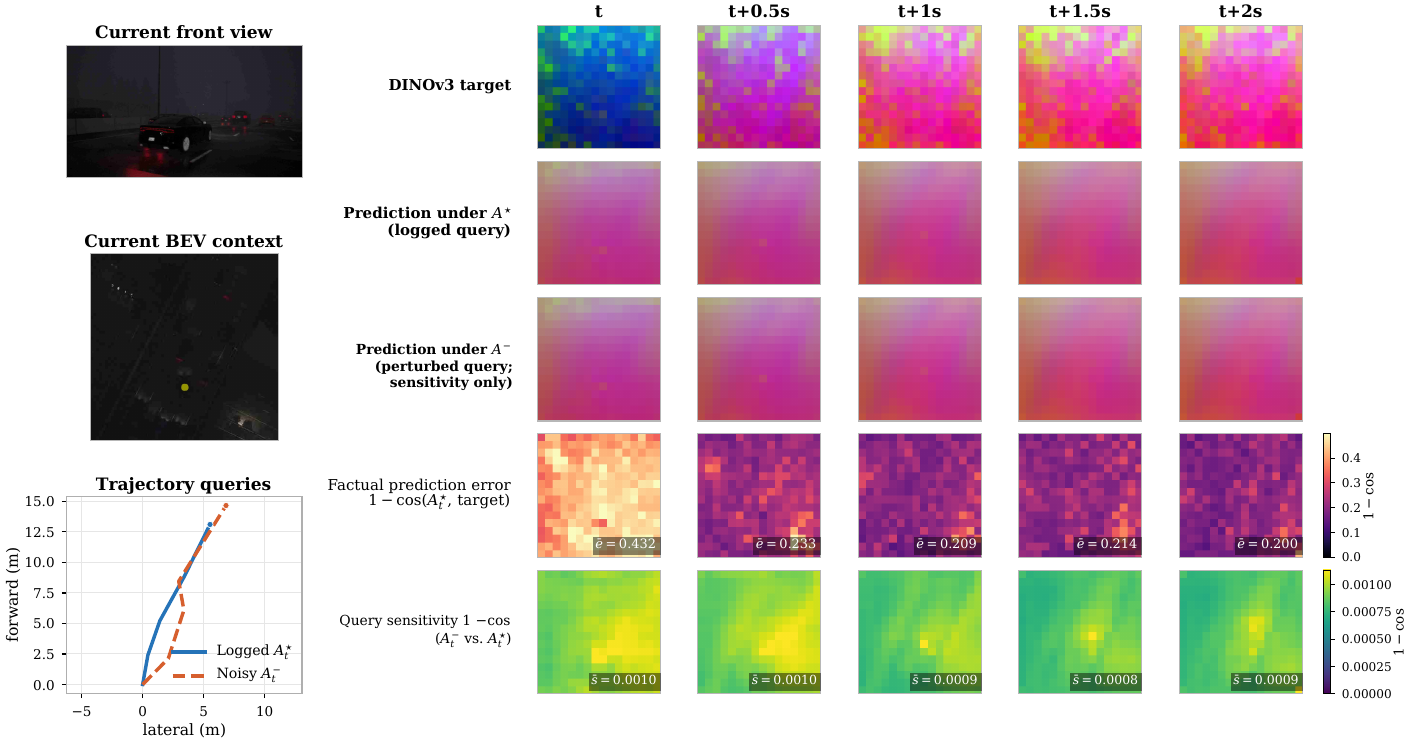}
\caption{Future-feature predictions under lateral coordinate perturbations. \emph{Top:} nighttime example from \texttt{Accident\_Town03\_Route102\_Weather20}, frame 99, perturbation seed 2033. \emph{Bottom:} rainy example from \texttt{Accident\_Town04\_Route205\_Weather23}, frame 163, seed 2040. The columns, rows, and visualization procedure follow Figure~\ref{fig:future_world_deceleration}.}
		\label{fig:future_world_lateral}
	\end{figure*}

Figures~\ref{fig:future_world_deceleration} and~\ref{fig:future_world_lateral} visualize \method{}'s feature predictions and trajectory-query responses on logged training examples. Each panel spans $t$ through $t+2\,$s in $0.5\,$s increments. From top to bottom, the rows show the frozen-DINOv3 target $Y^\star_{t+\tau}$, the prediction under the logged four-waypoint query $A^\star_t$, the prediction under the perturbed query $A^-_t$, the token-wise factual error $1-\cos(\widehat Y_{t+\tau}(A^\star_t),Y^\star_{t+\tau})$, and the query response $1-\cos(\widehat Y_{t+\tau}(A^-_t),\widehat Y_{t+\tau}(A^\star_t))$, where $\tau\in\{0,0.5,1,1.5,2\}\,$s. The final row measures feature sensitivity to the trajectory query.

Within each panel, a single PCA basis and color range are fitted jointly to the five factual DINOv3 targets and applied unchanged to the targets and both prediction sequences. The underlying tensors have shape $16\!\times\!16\!\times\!1024$. Factual error and query response are computed in the native 1024-dimensional feature space before PCA; nearest-neighbor enlargement is used for display.

Examples cover adverse weather, nighttime driving, and both perturbation types. For each frame and perturbation type, seeds 2027--2042 generate 16 candidate queries. Candidates are ranked by normalized feature-response RMS: the RMS difference between perturbed-query and logged-query predictions divided by the logged-query prediction RMS (denominator floor $10^{-12}$). Both RMS values average all five horizons, 256 spatial tokens, and 1024 channels before PCA. The displayed candidate is closest to the within-frame 75th percentile of this score. Frames were selected from a nine-frame pool ranked by cross-seed median scores. Table~\ref{tab:dino_horizon} reports aggregate prediction metrics on the separately collected evaluation episodes described in Appendix~\ref{app:dino_metrics}.

	\section{Closed-Loop Failure Analysis}
	\label{app:failure_analysis}

	\begin{table*}[t]
		\centering
		\caption{Mutually exclusive outcomes on all 220 routes. The four termination reasons partition the 30 incomplete routes.}
		\label{tab:failure_analysis}
		\begin{tabular}{@{}lcc@{}}
			\toprule
			Outcome & Routes & Fraction \\
			\midrule
			Strict success & 166 & 75.45\% \\
			Completed with disqualifying infraction & 24 & 10.91\% \\
			Did not complete & 30 & 13.64\% \\
			\quad Route deviation & 12 & 5.45\% \\
			\quad Route timeout & 12 & 5.45\% \\
			\quad Vehicle blocked & 5 & 2.27\% \\
			\quad Agent crash & 1 & 0.45\% \\
			\bottomrule
		\end{tabular}
	\end{table*}

We evaluate all 220 official route records, each with a unique route identifier. A strict success requires status \texttt{Completed} or \texttt{Perfect} and no infraction outside the ignored minimum-speed category. Table~\ref{tab:failure_analysis} summarizes the route outcomes.

Of 190 destination-reaching routes, 24 fail strict success because of a non-minimum-speed infraction: nine route deviations, eight vehicle collisions, five outside-route-lane violations, one red-light violation, and one layout collision. The remaining 30 routes terminate as shown in Table~\ref{tab:failure_analysis}. Across all 54 non-success routes, the most frequent route-level signals are route deviation (21), vehicle collision (13), and timeout (12). These signal counts overlap because a route may contain several infractions. The Wilson 95\% interval for $166/220$ is 69.36--80.67\%, describing finite-route uncertainty for this sweep.

	\section{Checkpoint Footprint}
	\label{app:footprint}

	\begin{table*}[t]
		\centering
\caption{Stored checkpoint footprint and native execution path for \baseplanner{} and \method{}, using decimal units. Parameter counts and payloads include auxiliary tensors that are bypassed during native inference.}
		\label{tab:execution_cost}
		\begin{tabular}{@{}lcc@{}}
			\toprule
			Quantity & \shortstack{\baseplanner[\\]} & \fullmodel{} \\
			\midrule
			Parameters & 4.063B & 4.086B ($+0.554\%$) \\
			Payload & 8.755GB & 8.800GB ($+45.04$MB) \\
			Native path & Planner + PID & Planner + PID \\
			\bottomrule
		\end{tabular}
	\end{table*}

The stored \baseplanner{} and \method{} checkpoints contain 4,063,126,528 and 4,085,644,803 parameters, respectively, an increase of 22.52M. Table~\ref{tab:execution_cost} reports their storage footprint and native execution paths.

	\section{Additional Evaluation Details}
	\label{app:evaluation_details}

	\subsection{DINOv3 Feature-Prediction Protocol}
\label{app:dino_metrics}

\paragraph{Episodes and sample selection.}
We collect new episodes with the frozen public DeepSight policy in CARLA 0.9.16. Five development routes are separate from the 15 evaluation routes, which span nine maps. For each evaluation route, a fixed quantile-based rule selects up to 25 valid windows from synchronized observations, including valid windows from failed episodes. Fourteen routes provide 25 windows and one provides 24, yielding $N=374$. These episodes are reserved for evaluation, with checkpoint weights and hyperparameters fixed throughout collection and scoring.

\paragraph{Models and references.}
The external-reference comparison evaluates one checkpoint each for \method{}, public DeepSight, and an external additive reference (B). The 220-route Phase-II comparisons use our in-house \baseplanner{}; public DeepSight supplies an additional pretrained reference. B was independently fine-tuned from public DeepSight for one epoch using DINO and RGB factual losses and RGB separation, with the RGB-to-planner gradient detached. These models provide prediction references with different training histories. For each model, a persistence baseline repeats its predicted current-time features at every future horizon. Repeating the teacher features from the actual current BEV adds a persistence reference with privileged BEV input.

\paragraph{Targets and metrics.}
Future BEV targets follow the executed poses and are aligned to the anchor ego frame. All models use identical target tensors from the same frozen DINOv3 teacher. Diagnostic metrics use the $K=256$ spatial patch tokens, each with $C=1024$ channels; training additionally supervises the CLS token. Predictions and targets are converted to FP32 for scoring, with MSE computed on raw feature values. For sample $n$ and offset $\tau$, define
\begin{equation}
\begin{aligned}
 m_{n,\tau}
 &=\frac{1}{KC}\sum_{i=1}^{K}\sum_{c=1}^{C}
 \left(\widehat Y_{n,\tau,i,c}-Y^\star_{n,\tau,i,c}\right)^2,\\
 c_{n,\tau}
 &=\frac{1}{K}\sum_{i=1}^{K}
 \frac{\widehat Y_{n,\tau,i}^{\top}Y^\star_{n,\tau,i}}
 {\max\!\left(\|\widehat Y_{n,\tau,i}\|_2,\epsilon\right)
  \max\!\left(\|Y^\star_{n,\tau,i}\|_2,\epsilon\right)},
 \\ &\qquad \epsilon=10^{-8}.
\end{aligned}
\label{eq:dino_eval_metrics}
\end{equation}
Let $\mathcal H=\{0.5,1.0,1.5,2.0\}\,\mathrm{s}$ and let $\mathcal N_r$ contain the selected windows of route $r$. The reported future MSE is
\begin{equation}
 \operatorname{MSE}_{\mathrm{future}}
 =\frac{1}{15}\sum_{r=1}^{15}\frac{1}{|\mathcal N_r|}
 \sum_{n\in\mathcal N_r}\frac{1}{4}\sum_{\tau\in\mathcal H}m_{n,\tau}.
\end{equation}
Future cosine uses the same aggregation with $c_{n,\tau}$. The four offsets in $\mathcal H$ define the future mean, and current-time metrics are reported separately. All scores are computed before PCA or image enlargement.

\begin{table*}[t]
\centering
\caption{Single-checkpoint external-reference comparison on all 374 evaluation windows from 15 newly collected episodes. Metrics average the four future offsets and then routes equally. The public and additive models are prediction references with different training histories. The last row uses privileged actual current-BEV input.}
\label{tab:dino_horizon}
\small
\begin{tabular}{@{}lcc@{}}
\toprule
Model or reference & Future MSE $\downarrow$ & Cosine $\uparrow$ \\
\midrule
Public DeepSight & 0.02329 & 0.82575 \\
\method{} & 0.03029 & 0.76771 \\
External additive reference (B) & 0.02340 & 0.82475 \\
\midrule
Public DeepSight: repeat predicted $t$ & 0.02463 & 0.81594 \\
\method{}: repeat predicted $t$ & 0.03110 & 0.76099 \\
External additive reference: repeat predicted $t$ & 0.02471 & 0.81506 \\
Repeat actual teacher $t$ (privileged) & 0.01635 & 0.89266 \\
\bottomrule
\end{tabular}
\end{table*}

\begin{table*}[t]
\centering
\caption{Single-checkpoint per-offset prediction for the external-reference comparison on the same 374 windows. Each entry averages windows within a route and then the 15 routes equally; Table~\ref{tab:dino_horizon} averages the four future columns.}
\label{tab:dino_per_horizon}
\small
\begin{tabular}{@{}llccccc@{}}
\toprule
Model & Metric & $t$ & $t{+}0.5$ & $t{+}1.0$ & $t{+}1.5$ & $t{+}2.0$ \\
\midrule
Public DeepSight & MSE $\downarrow$ & 0.02378 & 0.02330 & 0.02317 & 0.02335 & 0.02333 \\
                 & Cosine $\uparrow$ & 0.82259 & 0.82600 & 0.82661 & 0.82531 & 0.82505 \\
\method{}       & MSE $\downarrow$ & 0.03042 & 0.03025 & 0.03005 & 0.03024 & 0.03062 \\
                 & Cosine $\uparrow$ & 0.76726 & 0.76843 & 0.76949 & 0.76784 & 0.76509 \\
External additive reference & MSE $\downarrow$ & 0.02382 & 0.02337 & 0.02327 & 0.02348 & 0.02346 \\
                 & Cosine $\uparrow$ & 0.82208 & 0.82529 & 0.82565 & 0.82413 & 0.82393 \\
\bottomrule
\end{tabular}
\end{table*}

\paragraph{Single-checkpoint paired trajectory queries.}
For each recipient window, the alternative query is drawn from a different evaluation route, with speed difference at most $5\,\mathrm{m/s}$, endpoint-direction difference at most $20^\circ$, and a distinct waypoint sequence. The fixed random matching procedure finds an eligible donor for 371 windows. Factual-prediction metrics use all 374 windows; the paired-query analysis uses the 371 matched windows. Both queries are scored against the recipient's factual target, measuring the benefit of pairing that target with its executed trajectory. We first average paired differences within each route, then obtain 95\% percentile intervals from 10,000 route resamples with seed 20261109, keeping each route's windows together.

For the single \method{} checkpoint in Table~\ref{tab:dino_horizon}, route-equal factual and mismatched MSEs are 0.030293741 and 0.030349705, respectively: a relative increase of 0.185\%. The mean paired difference is $5.5964\times10^{-5}$, with a 95\% interval of $[0.3412,12.5433]\times10^{-5}$. The matched two-seed comparison appears in Appendix~\ref{app:matched_prediction}.

For the external additive reference, factual and mismatched-query MSEs on the same 371 windows are 0.023404210 and 0.023402824. Its paired difference is $-1.3866\times10^{-6}$, with a 95\% route-bootstrap interval of $[-3.3161\times10^{-6},2.0289\times10^{-7}]$, which spans zero.

\begin{figure*}[t]
\centering
\includegraphics[width=0.60\textwidth]{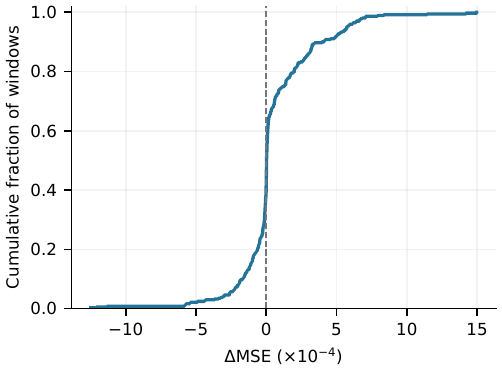}
\caption{Single-checkpoint trajectory-pairing effects for the \method{} checkpoint in Table~\ref{tab:dino_horizon}, across all 371 matched evaluation windows. Each window receives equal weight in the empirical cumulative distribution. We define $\Delta\mathrm{MSE}=\mathrm{MSE}_{\mathrm{mismatched}}-\mathrm{MSE}_{\mathrm{executed}}$, averaged over the four future offsets from $0.5$ to $2.0\,\mathrm{s}$. The dashed line marks zero; positive values indicate lower error with the executed query.}
\label{fig:paired_windows}
\end{figure*}

\subsection{Fixed-Query Responses and Readout Precision}
\label{app:query_precision}

Using the single-checkpoint \method{} and external additive reference from Appendix~\ref{app:dino_metrics}, we apply the same trajectory pair across scenes to measure how the query response depends on scene content. One pair is selected from development routes using trajectory metadata before inspecting predictions. Both queries must be within $5\,\mathrm{m/s}$ in speed and $20^{\circ}$ in endpoint direction of an evaluation window's executed query. This gives 252 eligible windows from 15 evaluation routes. Response energy and cross-scene variance use spatial DINO patches, with equal weight per route and equal weight per window within each route. Future values average the four horizons from $0.5$ to $2.0$ seconds; the current-time output at $t=0$ is measured separately.

For fixed queries $a_0,a_1$, the FiLM-conditioned linear readout has the analytic difference
\begin{equation}
\Delta(x)=W\left[(\gamma(a_1)-\gamma(a_0))\odot h(x)
+\beta(a_1)-\beta(a_0)\right],
\label{eq:fixed_query_delta}
\end{equation}
where $h(x)$ is the scene-dependent query state, $W$ is the DINO projection, and $\gamma=1+\Delta\gamma$ follows Equation~\ref{eq:latent_film}. FiLM couples trajectory modulation to $h(x)$; bias-only conditioning gives a scene-independent analytic difference. We evaluate native BF16 predictions and an FP32 readout replay. The replay casts the native hidden states, conditioning outputs, and loaded weights to FP32 for modulation and projection, retaining the same backbone outputs. Response energy is the mean squared difference, and variance is the mean squared deviation from its route-balanced cross-scene mean.

\begin{table}[t]
\centering
\caption{Single-checkpoint responses to the same trajectory pair across 252 windows, using the external-reference models from Appendix~\ref{app:dino_metrics}. All entries are in units of $10^{-6}$. F is TrajWAM; B is the external additive reference. The FP32 rows use the analytic readout difference.}
\label{tab:fixed_query_precision}
\resizebox{\columnwidth}{!}{
\begin{tabular}{llrrrr}
\toprule
 & & \multicolumn{2}{c}{Future} & \multicolumn{2}{c}{$t=0$} \\
\cmidrule(lr){3-4}\cmidrule(lr){5-6}
Model & Arithmetic & Energy & Variance & Energy & Variance \\
\midrule
F & Native BF16 & 7.2099 & 4.6981 & 6.7815 & 4.4884 \\
F & FP32 replay & 6.6529 & 4.1715 & 6.2456 & 3.9822 \\
B & Native BF16 & 0.9000 & 0.3505 & 0.9000 & 0.3517 \\
B & FP32 replay & 0.5221 & 0 & 0.5221 & 0 \\
\bottomrule
\end{tabular}}
\end{table}

TrajWAM produces scene-dependent query responses with FP32 readout arithmetic: its future cross-scene variance is $4.171457\!\times\!10^{-6}$, while the additive reference has zero analytic variance. This behavior follows the scene--trajectory interaction term in Equation~\ref{eq:fixed_query_delta}. The current-time and future columns in Table~\ref{tab:fixed_query_precision} show that this interaction operates across the supervised sequence.

\subsection{Native Planner Microbenchmark}
\label{app:native_profile}

We profile TrajWAM and the public DeepSight checkpoint on one NVIDIA GeForce RTX 5090 using BF16, batch size one, and SDPA, with PyTorch 2.7.1+cu128 and Transformers 4.56.0. Both models receive the same four fixed observations, one from each of four development routes, and use the same native preprocessing and generation configuration. Each model receives one warm-up generation followed by three timed generations per observation, yielding 12 measurements per model. CUDA synchronization is used for timing. Generation includes autoregressive reasoning and waypoint-token generation; file-to-waypoints timing additionally includes image preprocessing and waypoint decoding. PID control is measured separately. These measurements exclude simulator execution, sensor capture, file writing, model loading, and warm-up.

\begin{table}[t]
\centering
\caption{Native planner microbenchmark. Times are medians over 12 measurements; memory is the peak allocated GPU memory. The DeepSight reference is its public checkpoint.}
\label{tab:native_microbenchmark}
\small
\begin{tabular}{lrr}
\toprule
Measurement & DeepSight & TrajWAM \\
\midrule
Generation (s) & 3.4926 & 3.5086 \\
File to waypoints (s) & 3.7539 & 3.7676 \\
PID (ms) & 0.1940 & 0.1982 \\
Peak allocated memory (GiB) & 8.1813 & 8.2217 \\
Mean output tokens & 96.25 & 93.25 \\
\bottomrule
\end{tabular}
\end{table}

Both native prompts retain 1,305 future-query tokens. Native inference bypasses the DINO teacher, DINO readout, ActionEncoder, FiLM, and auxiliary decoders. Auxiliary tensors remain resident and contribute to the reported memory footprint. All timed runs use the uninstrumented execution path. Table~\ref{tab:native_microbenchmark} reports generation and file-to-waypoint latency, PID time, peak allocated GPU memory, and mean output length for the four observations.

\section{Multi-Seed Closed-Loop Evaluation on a Stratified Subset}
\label{app:subset120}
\begin{table*}[t]
		\centering
		\caption{Schedule-matched Phase-II comparison on all 220 Bench2Drive routes. Models and objectives follow Section~\ref{sec:experiments}.}
		\label{tab:matched_control}
		\begin{tabular}{@{}lcc@{}}
			\toprule
			Model & DS$\uparrow$ &  SR (\%)$\uparrow$ \\
			\midrule
			\baseplanner{} & 85.94 & 70.91\% \\
			\planningft{} & 86.22 & 71.36\% \\
			\rowcolor{oursblue}
			\fullmodel{} & \textbf{91.53} & \textbf{75.45\%} \\
			\bottomrule
		\end{tabular}
	\end{table*}

\paragraph{Route selection.}
We evaluate a fixed subset of 120 routes from the same Bench2Drive 220-route set, which contains 44 scenario types with five routes each. Scenario types are ordered by SHA256(seed|quota|scenario\_type), using selection seed 20261117. The first 32 types contribute three routes each and the remaining 12 contribute two, yielding $32\times3+12\times2=120$. Within these quotas, 1,024 candidate subsets are generated. Eligible candidates cover all 12 source maps and all starting-weather groups. The selected candidate minimizes the combined deviation of map and weather-group proportions from the full set; ties are broken by candidate index. Selection is determined entirely by route metadata. Scenario types define the strata; the overlapping driving-ability categories are reported separately in the full-set evaluation.

Weather groups use initial sun altitude: night below $0^\circ$, dawn/dusk from $0^\circ$ through $10^\circ$, and day above $10^\circ$. An adverse-weather tag denotes precipitation above zero, fog density at least 50, or wetness at least 50. The full-set/subset counts are 72/39 for day with adverse weather, 87/48 for other day weather, 44/24 for night with adverse weather, 9/5 for other night weather, and 8/4 for dawn/dusk with adverse weather.

\paragraph{Matched training variants.}
All variants use the same Phase-I initialization, training data, 24-epoch budget, optimizer, and loss weights. Planning-only FT uses $\mathcal L_{\mathrm{rec}}$; unconditioned future FT adds $2\mathcal L_{\mathrm{fact}}$ without a trajectory query; \method{} w/o separation uses trajectory-conditioned factual reconstruction. The additive variant and full \method{} additionally use $0.5\mathcal L_{\mathrm{sep}}$. The additive variant replaces FiLM's scale-and-shift modulation with additive conditioning. All five variants use training seeds 42 and 43. Each checkpoint, including both seeds of the shuffled-pairing control, is evaluated on the same 120 routes. The matched additive variant follows this common protocol in both closed-loop and prediction comparisons; the external additive reference in Appendix~\ref{app:dino_metrics} follows the separately stated training history.

\begin{table*}[t]
\centering
\caption{Individual training-seed results on the same 120 routes. SR values and their paired differences are rounded after aggregation.}
\label{tab:subset120_seeds}
\begin{tabular}{@{}lccc@{}}
\toprule
Variant & Training seed & DS $\uparrow$ & SR (\%) $\uparrow$ \\
\midrule
Planning-only FT & 42 & 82.64 & 64.17 \\
Planning-only FT & 43 & 83.16 & 65.00 \\
Unconditioned future FT & 42 & 82.11 & 65.83 \\
Unconditioned future FT & 43 & 82.49 & 66.67 \\
\method{} w/o separation & 42 & 86.28 & 66.67 \\
\method{} w/o separation & 43 & 86.52 & 67.50 \\
\method{} (additive) & 42 & 86.07 & 66.67 \\
\method{} (additive) & 43 & 85.83 & 65.83 \\
\method{} (correct pairing) & 42 & \textbf{88.69} & \textbf{69.17} \\
\method{} (correct pairing) & 43 & 88.21 & 68.33 \\
\bottomrule
\end{tabular}
\end{table*}

\paragraph{Paired component gains.}
Comparisons use shared training seeds and route identities. Across seeds 42 and 43, the complete objective improves DS by 5.55 points and SR by 4.17 percentage points over planning-only FT. Separation adds 2.05 DS points and 1.67 SR points over the trajectory-conditioned variant without separation. Across both seeds, trajectory conditioning adds 4.10 DS points and 0.83 SR points over unconditioned future supervision; full FiLM improves on additive conditioning by 2.50 DS points and 2.50 SR points. The complete method improves on planning-only FT by 6.05 and 5.05 DS points at seeds 42 and 43, respectively. Its corresponding gains over the no-separation variant are 2.41 and 1.69 DS points. Thus, both evaluated training seeds preserve the ordering of planning-only FT, trajectory-conditioned factual reconstruction, and the complete objective.

\paragraph{Conditional paired intervals.}
We use a scenario-stratified paired percentile bootstrap with 10,000 resamples and random seed 20261118. Within each of the 44 fixed scenario types, routes are sampled with replacement while retaining the original quota of two or three. Each resample uses identical route indices for both models and all common training seeds. For comparisons using two seeds, paired route differences are averaged over seeds before resampling. Training seeds and scenario categories remain fixed. The 2.5th and 97.5th percentiles give 95\% intervals conditional on these checkpoints and scenario strata. SR differences use $100$ times the difference of strict-success indicators. No finite-population correction is applied. Per-seed results separately describe the observed training-seed variation.

\begin{table*}[t]
\centering
\caption{Paired gains on the 120-route subset. SR differences are percentage points. Intervals are conditional, scenario-stratified paired route-bootstrap intervals; each comparison uses only its listed common training seeds.}
\label{tab:subset120_paired}
\begin{tabular}{@{}lccccc@{}}
\toprule
Comparison & Seeds & Mean $\Delta$DS & 95\% CI & Mean $\Delta$SR & 95\% CI \\
\midrule
Full $-$ planning-only FT & 42, 43 & $+5.55$ & $[+3.62,+7.50]$ & $+4.17$ & $[-0.83,+9.17]$ \\
Full $-$ no separation & 42, 43 & $+2.05$ & $[+0.54,+3.54]$ & $+1.67$ & $[-2.92,+6.25]$ \\
No separation $-$ unconditioned & 42, 43 & $+4.10$ & $[+2.46,+5.81]$ & $+0.83$ & $[-2.92,+5.00]$ \\
Full $-$ additive & 42, 43 & $+2.50$ & $[+1.03,+3.99]$ & $+2.50$ & $[-1.67,+7.08]$ \\
\bottomrule
\end{tabular}
\end{table*}

Across the four comparisons in Table~\ref{tab:subset120_paired}, mean DS and SR gains are positive. The conditional DS intervals are entirely above zero, while all SR intervals include zero.

\paragraph{Scenario-balanced aggregation.}
Let $\overline{\Delta\mathrm{DS}}_r$ be the paired DS difference on route $r$, averaged over the common training seeds, and let $n_h$ be the number of selected routes of scenario type $h$. We compute
\begin{equation}
\Delta\mathrm{DS}_{\mathrm{scenario}}=
\frac{1}{44}\sum_{h=1}^{44}\frac{1}{n_h}\sum_{r\in h}\overline{\Delta\mathrm{DS}}_r.
\end{equation}
Scenario-equal aggregation gives every scenario type the same weight. Route-equal aggregation averages the 120 paired route differences uniformly. Both statistics use the same stratified bootstrap draws, with their respective weights applied in each resample. Table~\ref{tab:subset120_scenarios} compares the two aggregations. The complete objective improves over planning-only FT in 24 of 44 scenario types, ties in 16, and declines in four, with a scenario-equal gain of 5.55 DS points. Removing each scenario type in turn preserves positive gains for the seed scopes listed in Table~\ref{tab:subset120_leaveout}.

\begin{table*}[t]
\centering
\caption{Scenario-balanced analysis on the fixed subset. The first two rows average seeds 42 and 43; the last two use seed 42. Sign counts list positive, tied, and negative scenario-level mean DS differences. Intervals use the same conditional stratified route resampling with equal scenario weights.}
\label{tab:subset120_scenarios}
\begin{tabular}{@{}lccccc@{}}
\toprule
Comparison & Seeds & Route-equal $\Delta$DS & Scenario-equal $\Delta$DS & 95\% CI & Signs ($+ / 0 / -$) \\
\midrule
Full $-$ planning-only FT & 42, 43 & $+5.55$ & $+5.55$ & $[+3.42,+7.68]$ & 24 / 16 / 4 \\
Full $-$ no separation & 42, 43 & $+2.05$ & $+2.10$ & $[+0.46,+3.75]$ & 22 / 19 / 3 \\
No separation $-$ unconditioned & 42 & $+4.17$ & $+4.07$ & $[+2.14,+6.08]$ & 21 / 20 / 3 \\
Full $-$ additive & 42 & $+2.62$ & $+2.76$ & $[+1.28,+4.28]$ & 20 / 20 / 4 \\
\bottomrule
\end{tabular}
\end{table*}

\begin{table*}[t]
\centering
\caption{Leave-one-scenario-type-out analysis. Each range spans the 44 recomputed mean DS differences after removing one scenario type and renormalizing the remaining weights. These are sensitivity ranges. The first two rows average seeds 42 and 43; the last two use seed 42.}
\label{tab:subset120_leaveout}
\begin{tabular}{@{}lcccc@{}}
\toprule
Comparison & Seeds & Full-subset $\Delta$DS & Route-equal range & Scenario-equal range \\
\midrule
Full $-$ planning-only FT & 42, 43 & $+5.55$ & $[+5.05,+5.93]$ & $[+5.09,+6.02]$ \\
Full $-$ no separation & 42, 43 & $+2.05$ & $[+1.65,+2.60]$ & $[+1.69,+2.60]$ \\
No separation $-$ unconditioned & 42 & $+4.17$ & $[+3.77,+4.56]$ & $[+3.53,+4.52]$ \\
Full $-$ additive & 42 & $+2.62$ & $[+2.15,+3.06]$ & $[+2.34,+3.16]$ \\
\bottomrule
\end{tabular}
\end{table*}

\subsection{Training-Pairing Control}
\label{app:training_pairing}
The within-group shuffled-pairing control uses the complete architecture, with the same initialization, training budget, and objectives. It disrupts the association between trajectory queries and their recipient scenes while preserving planning supervision and recipient factual targets. Table~\ref{tab:training_pairing} reports the two training seeds and their matched comparisons on the fixed 120 routes. Correct pairing improves DS by 4.67 and 3.83 at seeds 42 and 43, respectively. The mean gain is 4.25 DS; the conditional paired interval is $[2.64,5.89]$. The mean SR gain is 1.67 percentage points, with interval $[-2.50,5.84]$.

\begin{table*}[t]
\centering
\caption{Correct versus within-group shuffled training pairing on the fixed 120 routes. Both variants use seeds 42 and 43. SR and its differences are percentages and percentage points, respectively. The last row reports paired differences and conditional scenario-stratified route-bootstrap 95\% intervals.}
\label{tab:training_pairing}
\begin{tabular}{@{}lcccc@{}}
\toprule
Training / comparison & Seed 42: DS / SR & Seed 43: DS / SR & Mean DS $\pm$ SD & Mean SR $\pm$ SD \\
\midrule
Shuffled pairing & 84.02 / 66.67 & 84.38 / 67.50 & $84.20\pm0.25$ & $67.08\pm0.59$ \\
\method{} (correct pairing) & 88.69 / 69.17 & 88.21 / 68.33 & $88.45\pm0.34$ & $68.75\pm0.59$ \\
\midrule Correct $-$ shuffled & +4.67 / +2.50 & +3.83 / +0.83 & $+4.25\ [2.64,5.89]$ & $+1.67\ [-2.50,5.84]$ \\
\bottomrule
\end{tabular}
\end{table*}

\subsection{Matched Planning--Prediction Diagnosis}
\label{app:matched_prediction}
Table~\ref{tab:matched_prediction} evaluates the checkpoints used in the two-seed closed-loop comparison. Future MSE and persistence use all 374 windows from Appendix~\ref{app:dino_metrics}. Persistence repeats each model's predicted current-time features at the four future offsets. Query benefits use the same 371 matched windows, recipient observations, targets, executed queries, and mismatched queries for every model and seed. At evaluation, the shuffled-training model also receives each recipient's executed query. Metrics average future offsets and windows within routes, then the 15 routes and two seeds equally.

\begin{table*}[t]
\centering
\caption{Matched two-seed planning--prediction diagnosis. DS is the mean on 120 closed-loop routes; future and persistence MSE use 374 windows. Query gains and their conditional 95\% intervals use 371 windows and are in units of $10^{-5}$. Unconditioned supervision has no trajectory-query comparison.}
\label{tab:matched_prediction}
\begin{tabular}{@{}lrrrrl@{}}
\toprule
Model & Mean DS $\uparrow$ & Future MSE $\downarrow$ & Persistence MSE $\downarrow$ & Query gain $\uparrow$ & 95\% CI \\
\midrule
Unconditioned future FT & 82.30 & 0.03110 & 0.03170 & --- & --- \\
\method{} w/o separation & 86.40 & 0.02985 & 0.03055 & +1.60 & $[-0.60,+4.00]$ \\
Matched additive & 85.95 & 0.03055 & 0.03115 & +0.70 & $[-1.00,+2.80]$ \\
\method{} & 88.45 & 0.03020 & 0.03101 & +6.00 & $[+0.80,+12.50]$ \\
\bottomrule
\end{tabular}
\end{table*}

Trajectory conditioning improves prediction quality, lowering future MSE by 4.02\% relative to unconditioned supervision. Separation adds planning gains: DS rises from 86.40 to 88.45, and future MSE increases from 0.02985 to 0.03020. The complete method improves both DS and future MSE over matched additive conditioning. Thus, the component controls associate conditioning with better reconstruction and planning, and separation with additional planning gains.

\paragraph{Paired query benefits.}
For model $m$, seed $s$, and route $r$, let $\delta_{m,s,r}$ be the mean of mismatched minus executed MSE over the route's matched windows and four future offsets. We average seeds within each route, $\bar\delta_{m,r}=(\delta_{m,42,r}+\delta_{m,43,r})/2$, then average the 15 routes equally. Each of 10,000 bootstrap replicates draws 15 routes with replacement, retaining all their windows and using identical route indices and multiplicities across models and seeds. Percentiles 2.5 and 97.5 give intervals conditional on the evaluated checkpoints. For correct minus shuffled training, each replicate averages the paired differences $\bar\delta_{\mathrm{correct},r}-\bar\delta_{\mathrm{shuffled},r}$ directly. Table~\ref{tab:training_query_benefit} and Figure~\ref{fig:paired_routes} report the results.

\begin{table*}[t]
\centering
\caption{Executed-query benefits after correct or shuffled training pairing. All gains and conditional paired route-bootstrap 95\% intervals are in units of $10^{-5}$ MSE, averaged over seeds 42 and 43.}
\label{tab:training_query_benefit}
\begin{tabular}{@{}lrr@{}}
\toprule
Training / comparison & Mean query gain & 95\% CI \\
\midrule
Shuffled pairing & +0.30 & $[-1.20,+1.80]$ \\
Correct pairing & +6.00 & $[+0.80,+12.50]$ \\
Correct $-$ shuffled & +5.70 & $[+0.50,+11.80]$ \\
\bottomrule
\end{tabular}
\end{table*}

\section{Limitations}
\label{sec:limitations}
Evaluation covers CARLA/Bench2Drive, with one training seed for the full set and two for all component and shuffled-pairing controls on the 120-route subset. Query diagnostics measure factual reconstruction and responses to alternative trajectories whose outcomes are unobserved; separation also acts on the current-time anchor. The public initializer's pretraining route coverage is unknown. Further evaluation can measure alternative-action outcomes, training-seed variation, and real-world transfer.

\section{Ethics and Reproducibility}
\subsection*{Ethics statement}
    All driving experiments use CARLA/Bench2Drive simulations.
    The study involves no human subjects or personal data.

	\subsection*{Reproducibility Statement}
    Section~\ref{sec:method} specifies the model and training objectives,
    Section~\ref{sec:experiments} describes the experimental setup, and
    Appendices~\ref{app:implementation} and~\ref{app:evaluation_details}
    provide implementation and evaluation details. The supplementary
    code archive provides the native model runtime, driving agent, PID controller,
    benchmark routes and scenarios, and scripts for multi-GPU closed-loop
    evaluation and metric aggregation. Its README and reproduction guide describe
    environment setup and evaluation using a complete trained TrajWAM checkpoint.
    The code archive does not include model weights.
    Appendix~\ref{app:dino_metrics} specifies the new-episode prediction protocol,
    including sample selection, target construction, and aggregation. Appendix~\ref{app:subset120} describes the stratified route subset and multi-seed comparisons.

\end{document}